\documentclass[letterpaper]{article} 
\usepackage{aaai2026}  
\usepackage{times}  
\usepackage{helvet}  
\usepackage{courier}  
\usepackage[hyphens]{url}  
\usepackage{graphicx} 
\usepackage{natbib}  
\usepackage{caption} 
\usepackage{algorithm}
\usepackage{algorithmic}
\usepackage{amsmath}
\usepackage{amssymb}
\usepackage{color}
\usepackage{booktabs}
\usepackage{multirow}
\usepackage{rotating} 
\usepackage{enumitem}
\usepackage{needspace}

\usepackage{newfloat}
\usepackage{listings}
\DeclareCaptionStyle{ruled}{labelfont=normalfont,labelsep=colon,strut=off} 
\floatstyle{ruled}
\newfloat{listing}{tb}{lst}{}
\floatname{listing}{Listing}
\title{MdaIF: Robust One-Stop Multi-Degradation-Aware Image Fusion with Language-Driven Semantics}
\author{
    Jing Li\textsuperscript{\rm 1,\rm 2}, 
    Yifan Wang\textsuperscript{\rm 3}, 
    Jiafeng Yan\textsuperscript{\rm 3}, 
    Renlong Zhang\textsuperscript{\rm 3}, 
    Bin Yang\textsuperscript{\rm 4}\thanks{Corresponding author}
}
\affiliations{
    \textsuperscript{\rm 1}Key Laboratory of Geographic Information Science (Ministry of Education), East China Normal University, Shanghai 200241, China\\
    \textsuperscript{\rm 2}Key Laboratory of Spatial-temporal Big Data Analysis and Application of Natural Resources in Megacities, Ministry of Natural Resources, East China Normal University, Shanghai 200241, China\\
    \textsuperscript{\rm 3}School of Information, Central University of Finance and Economics, Beijing 102206, China\\
    \textsuperscript{\rm 4}School of Artificial Intelligence and Robotics, Hunan University, Changsha 410082, China\\

    jingli@geoai.ecnu.edu.cn, \{2024212461, yan.jiafeng, 2023312325\}@email.cufe.edu.cn, binyang@hnu.edu.cn
}

\usepackage{bibentry}

\begin{document}

\maketitle

\begin{abstract}
Infrared and visible image fusion aims to integrate complementary multi-modal information into a single fused result. However, existing methods 1) fail to account for the degradation visible images under adverse weather conditions, thereby compromising fusion performance; and 2) rely on fixed network architectures, limiting their adaptability to diverse degradation scenarios. To address these issues, we propose a one-stop degradation-aware image fusion framework for multi-degradation scenarios driven by a large language model (MdaIF). Given the distinct scattering characteristics of different degradation scenarios (e.g., haze, rain, and snow) in atmospheric transmission, a mixture-of-experts (MoE) system is introduced to tackle image fusion across multiple degradation scenarios. To adaptively extract diverse weather-aware degradation knowledge and scene feature representations, collectively referred to as the semantic prior, we employ a pre-trained vision-language model (VLM) in our framework. Guided by the semantic prior, we propose degradation-aware channel attention module (DCAM), which employ degradation prototype decomposition to facilitate multi-modal feature interaction in channel domain. In addition, to achieve effective expert routing, the semantic prior and channel-domain modulated features are utilized to guide the MoE, enabling robust image fusion in complex degradation scenarios. Extensive experiments validate the effectiveness of our MdaIF, demonstrating superior performance over SOTA methods.
\end{abstract}

\begin{links}
    \link{Code}{https://github.com/doudou845133/MdaIF}
\end{links}

\section{Introduction}
Infrared and visible image fusion (IVF) aims to integrate the complementary advantages of multi-modal sensors~\cite{zongshu2021,zongshu2024}. Infrared sensor captures thermal radiation from objects, presenting thermodynamic characteristics via pixel intensity distributions but lacking fine-grained texture details. In contrast, visible sensor relies on surface reflectance to provide rich texture information. IVF significantly enhances scene perception and offers more robust inputs for downstream vision tasks such as assisted driving~\cite{bao2023heat} and intelligent surveillance~\cite{surveillance2018zhang}. However, in the imaging process, due to the wavelength difference between infrared and visible light, adverse weather conditions such as haze, rain, and snow cause significant scattering of visible light, whereas infrared imaging remains relatively unaffected, especially over short distances~\cite{infraredgood,infraredanalysis}. As a result, texture details in visible images are poorly preserved, thereby impairing the performance of IVF and further affecting high-level vision tasks. Therefore, \emph{accounting for adverse weather degradations in IVF is crucial to enhancing its generalization and applications}.

\begin{figure}[t]
  \centering
  \includegraphics[width=1\columnwidth]{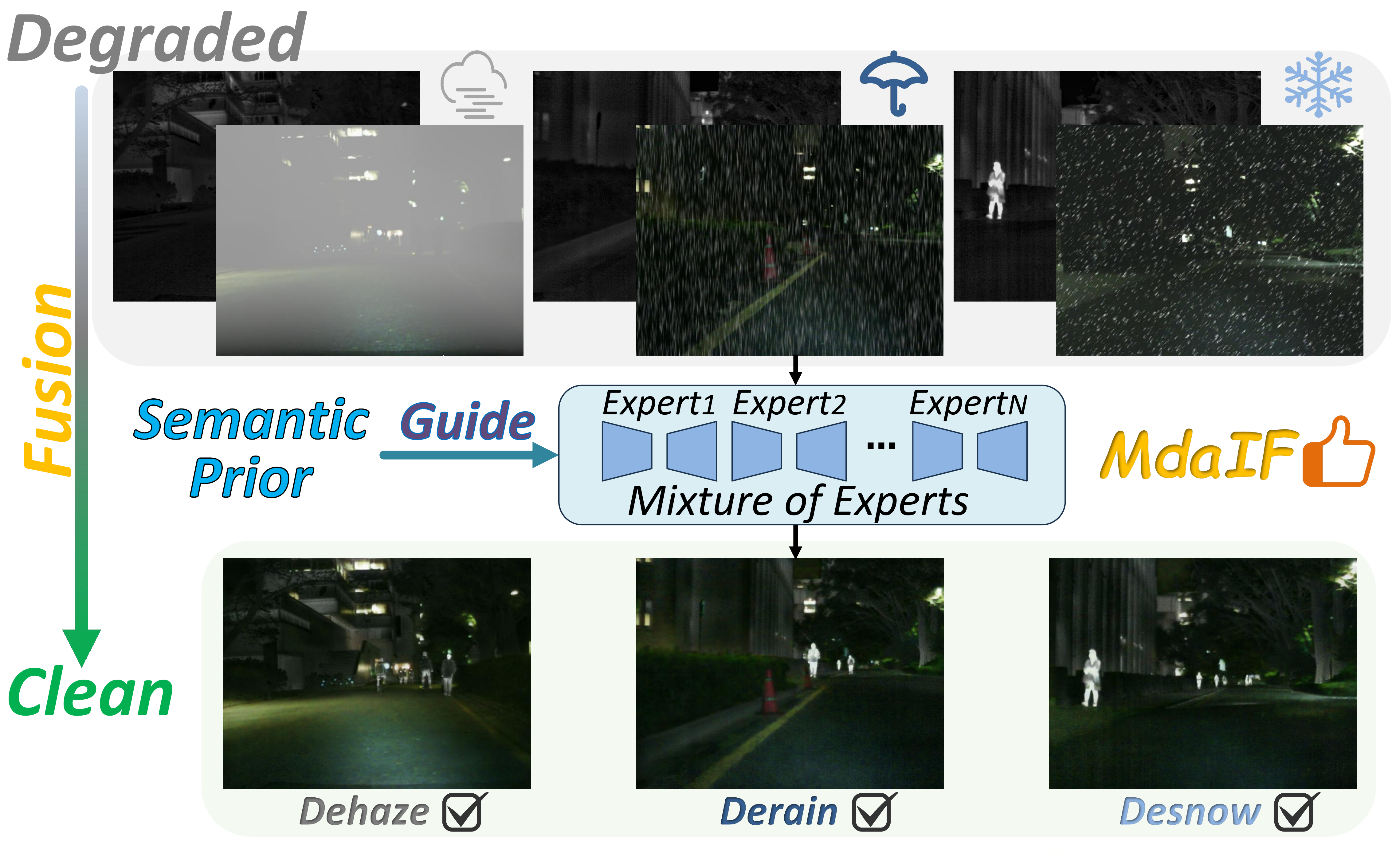}
  \small
  \caption{Our fusion results under haze, rain, and snow conditions, producing clean outputs from degraded inputs.}
    \label{fig:title}
\end{figure}
\begin{figure*}[t!]
  \centering
  \includegraphics[width=1\textwidth, keepaspectratio]{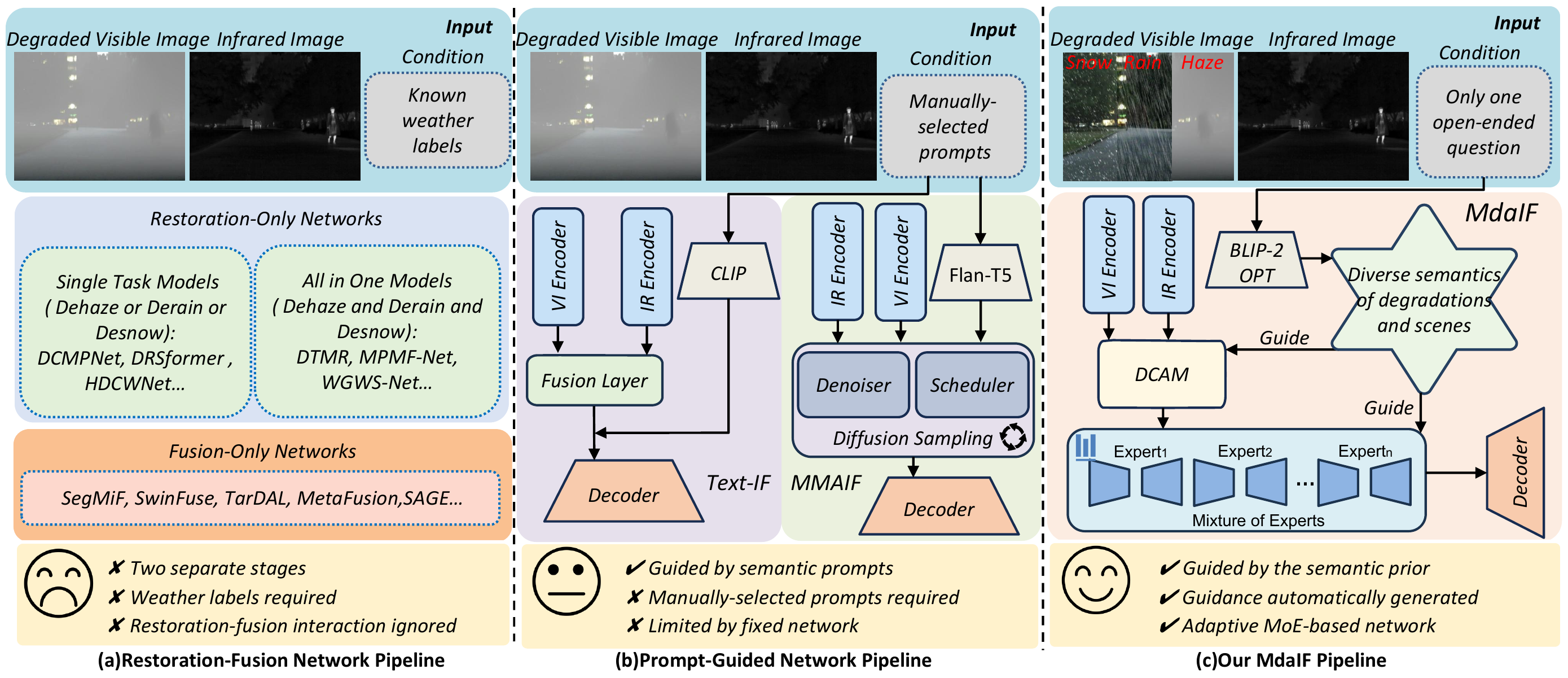}
  \small
  \caption{Comparisons with previous pipelines.}
    \label{fig:relatedwork}
\end{figure*}
To address the above issue, with the advancement of image restoration~\cite{yang2024language,restoration} and multi-modal image fusion~\cite{wu2025sage,zhao2024equivariant} technologies, a straightforward approach to infrared and degraded visible image fusion (IdVF) is to cascade image restoration and multi-modal fusion models, as Fig.~\ref{fig:relatedwork} (a), however, such a sequential design often leads to suboptimal feature alignment and error accumulation across tasks. In addition, we cascade SOTA image restoration and fusion models for IdVF, and experiments show this strategy is insufficient to address the above issue, as detailed in the experimental section. In contrast, Fig.~\ref{fig:title} shows that our method yields superior fusion results on degraded visible and infrared inputs.

IdVF has been gaining momentum as a prominent research focus, several methods have begun to incorporate VLMs or large language models (LLMs) to enhance performance in degraded image fusion scenarios. For example, Text-IF~\cite{textif2024text} leverages prompt-guided mechanisms built upon CLIP~\cite{CLIP2021learning}, a typical VLM, to facilitate image fusion in degradation scenarios like low illumination and overexposure, yet it overlooks the adverse weather-induced degradations that significantly affect fusion performance. In addition, MMAIF~\cite{mmaif_cao2025} combines diffusion model with Flan-T5~\cite{flant5_2024scaling}, a representative LLM, to perform image fusion under diverse degradations. However, both MMAIF and Text-IF rely on fixed prompts derived from ground-truth degradation types and employ a single unified network for all conditions, which limits the performance of IdVF, as depicted in Fig.~\ref{fig:relatedwork} (b). Accordingly, we question: \emph{can IdVF under diverse adverse weather conditions be adaptively achieved by leveraging multiple networks to generate clean fused image, without requiring relying on ground-truth degradation types?}

Therefore, we propose a one-stop degradation-aware image fusion framework for multi-degradation scenarios driven by VLM. \textbf{The motivations are shown as follow}:

1) \emph{The distinct atmospheric scattering characteristics of haze, rain, and snow hinder the ability of a fixed network to generalize across different degradation scenarios.} Specifically, the differences in particle characteristics—micron-sized droplets in haze, millimeter-scale raindrops in rain, and ice crystals in snow—lead to fundamentally different atmospheric scattering models. Fixed network architectures exhibit limited capacity in capturing the heterogeneous degradation patterns associated with various adverse weather conditions. For example, the transmission map, which is essential for haze removal~\cite{dehazeformer2023vision}, becomes ineffective when applied to deraining scenarios~\cite{fu2017clearing}.

To this end, we propose a MoE-based framework, where a set of specialized experts are tailored to effectively solve the IdVF task under multiple degradation conditions. To mitigate the performance degradation caused by imbalanced expert selection—where one expert is overloaded with multiple tasks while others remain inactive—we propose a semantic prior-guided expert routing strategy. The proposed strategy employs the semantic prior derived from the VLM to interact with the modulated features, establishing a task-specific expert routing mechanism. This enables the model to adaptively select and compose appropriate experts based on prior knowledge, thereby effectively addressing IdVF tasks under diverse degradation scenarios.

2) \emph{Existing IdVF methods rely on ground-truth degradation types as prompts, which limits their flexibility and constrains their applicability.} Specifically, relying on fixed prompts derived from ground-truth degradation types merely exploits the text-visual alignment capabilities of these models, without leveraging their potential to understand complex degradation scenarios, thereby limiting the flexibility of IdVF under diverse degradation conditions.
\begin{figure*}[t!]
  \centering
  \includegraphics[width=1\textwidth]{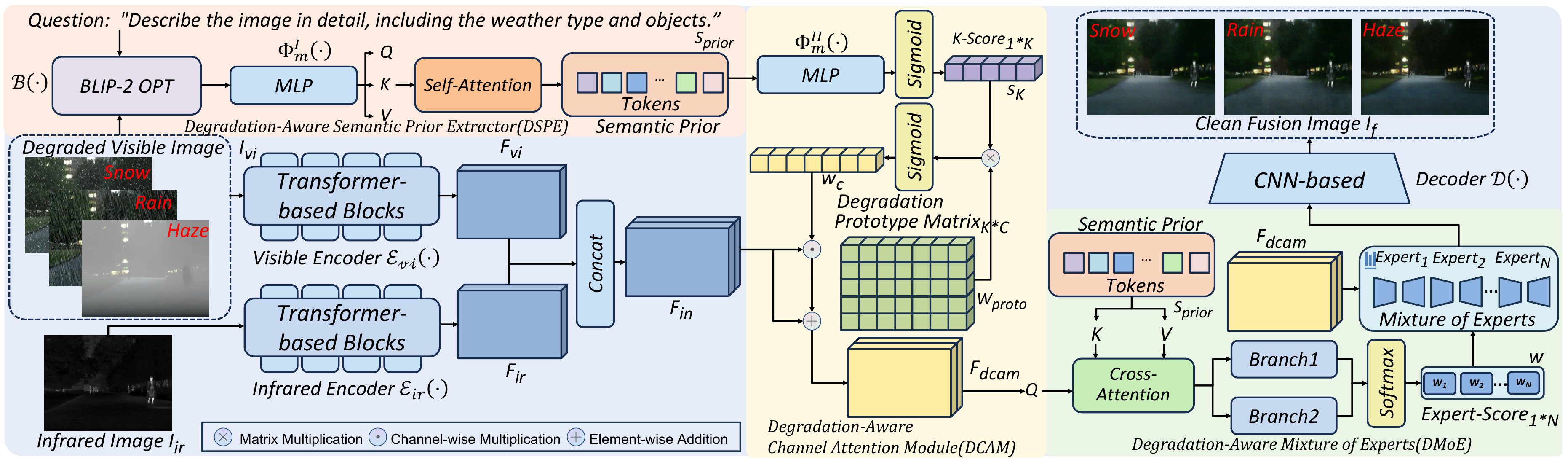}
  \small
  \caption{Overview of the proposed network architecture.}
    \label{fig:framework}
\end{figure*}

Therefore, we introduce the VLM into our method, which fully leverages the scene understanding capabilities—demonstrated effective in restoration tasks~\cite{yang2024language}—to enable the model to adaptively perceive degradation types, thereby guiding MoE to select appropriate task-specific experts. Rather than using the VLM solely as a classifier for degradation types, we further exploit its semantic understanding of the scene to enhance feature interaction and representation within the DCAM module.
Our contributions can be summarized as follows:
\begin{itemize}[itemsep=0pt, parsep=0pt, topsep=0pt]
\item Our method breaks the conventional reliance on sequential subtask processing by introducing joint optimization for degraded multi-modal fusion, thus alleviating feature misalignment and error accumulation.
\item To break the dependency on ground-truth degradation types and improve the model's adaptability and applicability, we propose a VLM-based adaptive degradation-aware framework for multi-degradation IdVF.
\item To address the limitations of fixed networks in modeling heterogeneous atmospheric degradations, we propose a MoE-based degradation-aware image fusion framework, which is guided by the semantic prior of the VLM and feature representations strengthened by DCAM.
\end{itemize}

\section{Related Work}
\subsection{General Image Fusion}
IVF has emerged as a critical subfield of computer vision and has received extensive attention in recent years~\cite{ma2019infrared,zhang2023visible}. With the advancement of algorithms, IVF has evolved from traditional manually designed fusion rules to end-to-end deep learning-based frameworks, which includes CNN- and GAN-based local feature fusion methods~\cite{ren2018infrared,li2019coupled}, Transformer-based global feature fusion methods~\cite{yangbin}, hybrid models that integrate local and global features~\cite{chen2023thfuse}, and diffusion-based denoising fusion frameworks~\cite{zhao2023ddfm}. In terms of fusion objectives, IVF has evolved from visually-oriented methods to task-driven fusion strategies designed to support downstream applications such as semantic segmentation~\cite{liu2023paif} and object detection~\cite{jiang2024m2fnet}. With respect to fusion conditions, IVF has developed from the fusion of well-registered source images to that of unregistered inputs~\cite{li2025mulfs}. However, existing methods rarely consider or simultaneously address multi-type degradation caused by diverse adverse weather scenarios.
\subsection{Prompt-Guided Image Fusion and Degradation-Awareness}
Driven by the rapid advancement of VLMs and LLMs, recent studies have integrated these models into IVF frameworks to enhance adaptability and flexibility. On one hand, such models offer improved capabilities for adaptive fusion by enabling fine-grained control over specific regions of interest. For example, TeRF~\cite{wang2024terf} employs text-driven and region-aware mechanisms to achieve semantically guided IVF. On the other hand, they have been utilized to facilitate degradation-aware IVF. For instance, TexIF leverages prompt-based guidance to perceive specific types of degradations and perform corresponding IdVF. Additionally, OmniFuse~\cite{zhang2025omnifuse} and MMAIF integrate semantic prompts with diffusion model to address degradation-aware fusion tasks, and OmniFuse supports prompt-based region-specific enhancement to improve the flexible of IVF.

However, they overlook the heterogeneous degradation factors caused by diverse adverse weather conditions. Furthermore, all the aforementioned methods rely on ground-truth degradation labels and fixed networks, without considering the distinct transmission patterns and mathematical formulations associated with different types of degradation. Therefore, we propose a VLM-based adaptive degradation-aware framework for multi-degradation IdVF.
\section{Method}
As illustrated in Fig.~\ref{fig:framework}, the MdaIF framework is fundamentally guided by semantic priors. In the following sections, we present a comprehensive analysis of our method from three key perspectives: problem formulation, network architecture, and loss functions.
\subsection{Problem Formulation} 
General IVF methods typically adopt a fixed network \(\theta_n\) and implicitly assume a clean visible image $\tilde{I_{vi}}$, without considering the effects of severe degradation. The network is designed to learn a predefined fusion function \(\mathcal{F}_{{if}}(\cdot)\) for generating the fusion result. The fusion process can be formulated as:
\begin{equation}
I_{f} = {\mathcal{F}_{{if}}}(\tilde{I_{vi}}, I_{ir}; \theta_n).
\end{equation}

However, in adverse weather, visible images often suffer severe degradation that fixed networks cannot handle effectively. Thus, we design an adaptive MoE-based IdVF network guided by semantic priors, which enable degradation-aware fusion without relying on ground-truth degradation labels. The process is defined as:
\begin{equation}
I_{f} = {\mathcal{F}_{mdaif}}(I_{vi}, I_{ir}, S_{prior};\theta_{moe}),
\end{equation}
where $I_{vi}$ denotes the degraded visible image, $S_{prior}$ represents the semantic prior derived from the VLM, $\mathcal{F}_{mdaif}(\cdot)$ is the multi-degradation-aware fusion function, and $\theta_{moe}$ indicates the parameters of the MoE-based network.
\subsection{Network Structure}
\subsubsection{Encoder}
In our fusion pipeline, the degraded visible image $ I_{vi} \in \mathbb{R}^{H \times W \times 3} $ and the infrared image $ I_{ir} \in \mathbb{R}^{H \times W \times 1} $ are independently encoded by encoders $ \mathcal{E}_{vi} $ and $ \mathcal{E}_{ir} $, which are configured as transformer-based structures following SegFormer~\cite{xie2021segformer}. Each encoder comprises 4 transformer blocks. The encoded features $F_{vi}$ and $F_{ir}$ are concatenated along the channel dimension and subsequently fed into DCAM, which is formulated as:
\begin{equation}
    F_{vi} = \mathcal{E}_{vi}(I_{vi}), F_{ir} = \mathcal{E}_{ir}(I_{ir}), 
    F_{in} = \mathrm{Cat}(F_{vi}, F_{ir}),
\end{equation}
where the operator $ \mathrm{Cat}(\cdot) $ denotes channel-wise concatenation, and $ F_{in} \in \mathbb{R}^{H \times W \times C} $, with $ C $ representing the number of channels.
\subsubsection{Degradation-Aware Semantic Prior Extractor (DSPE)}
We employ BLIP-2 OPT 2.7B~\cite{li2023blip}, a pre-trained VLM, to extract semantic prior from the degraded visible image. Specifically, we adopt a visual question answering (VQA) paradigm, where the degraded visible image ${I}_{vi}$ and an open-ended question prompt $\mathcal{P}_q$ are jointly fed into the VLM to generate degradation-aware original semantic prior, denoted as ${S_{org}} \in \mathbb{R}^{S \times C_{org}}$, where $S$ represents the sequence length and $C_{org}$ denotes the original token dimension. We formulate it as:
\begin{align}
    S_{org} = \mathcal{L}({\mathcal{B}}(I_{vi}, \mathcal{P}_q)),
\end{align}
where \(\mathcal{B}(\cdot)\) denotes the BLIP-2 OPT model, and \(\mathcal{L}(\cdot)\) represents the operator used to extract the features from its last hidden layer.

To better align the original semantic prior with the feature space of our framework, we refine it through an MLP $\Phi_{m}^I(\cdot)$ to reduce the embedding dimension, producing the compressed representation ${S}_{embed} \in \mathbb{R}^{S \times C}$:
\begin{align}
    {S}_{embed} = \mathcal{N}_{layer}(\Phi_{m}^{I}({S}_{org})),
\end{align}
where $\mathcal{N}_{layer}(\cdot)$ denotes layer normalization, which promotes stable training and faster convergence. Here, $C$ denotes the operational feature dimension in our framework.

To emphasize the most salient tokens, we apply a self-attention mechanism to adaptively reweight their importance in the semantic prior. This adaptive refinement produces a well-aligned and semantically enriched prior, denoted as ${S}_{prior}$, which provides a robust and informative guidance for DCAM and expert routing in DMoE. The self-attention operation is defined as follows:
\begin{align}
    {Q} = {S}_{embed} {W}_Q, {K} &= {S}_{embed} {W}_K, {V} = {S}_{embed} {W}_V,\\
    {S}_{prior} & = {softmax}\left(\frac{{Q}{K}^T}{\sqrt{d_k}}\right) {V},
\end{align}
where ${W}_Q, {W}_K, {W}_V \in \mathbb{R}^{C \times d_k}$ are learnable projection matrices for the query, key, value, and $d_k$ is the scaling factor.

\begin{figure}[t!]
  \centering
  \includegraphics[width=1\columnwidth]{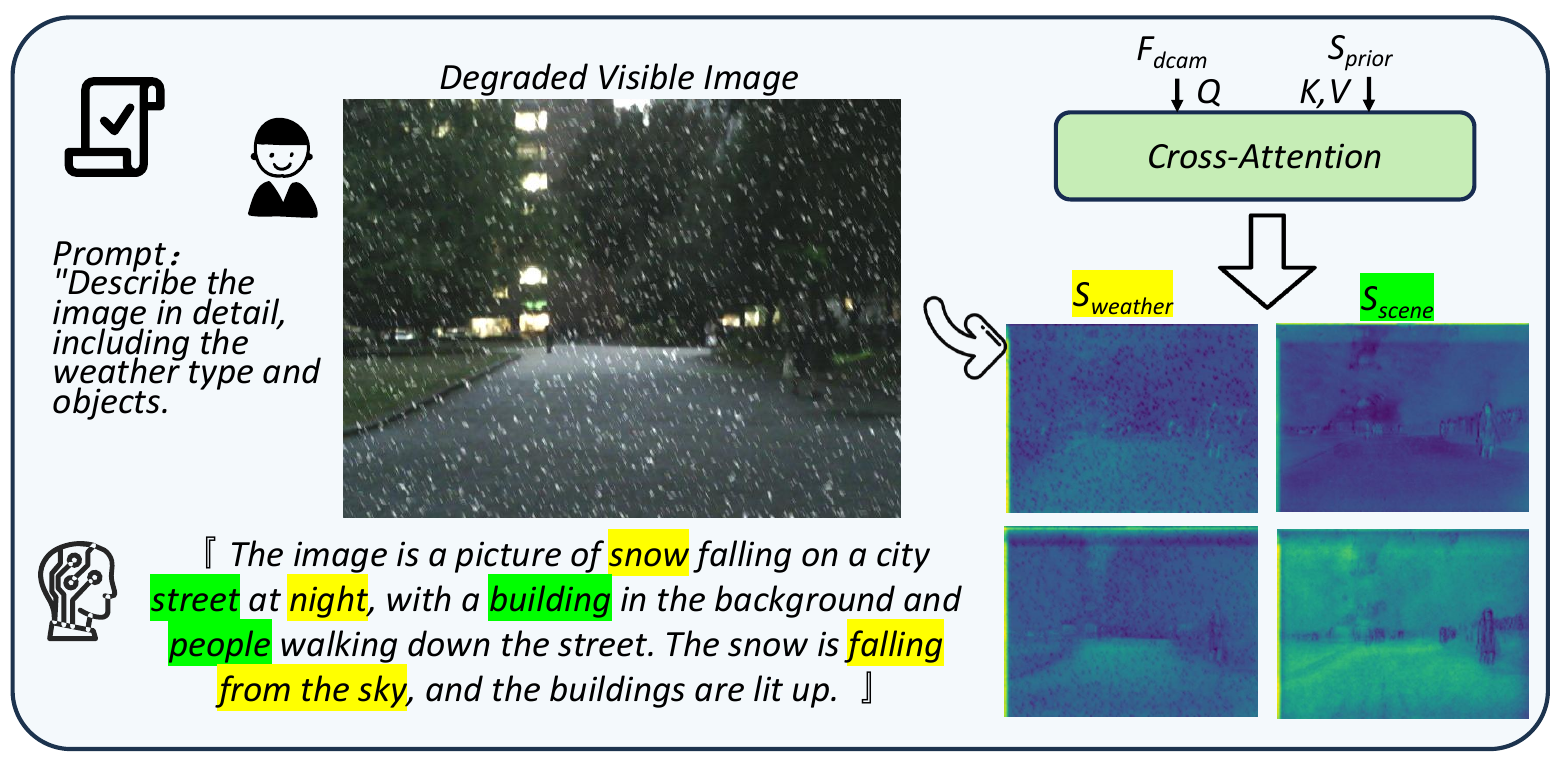}
  \small
  \caption{The process of semantic prior extraction and its deep interaction with image features.}
    \label{fig:VQA}
\end{figure}

An example of \textbf{the semantic prior extraction is illustrated in Fig.~\ref{fig:VQA}}, where the prior consists of two parts: $S_{{weather}}$, representing weather-aware degradation knowledge, and $S_{{scene}}$, capturing the scene features. In subsequent expert routing, image features modulated by DCAM interact with this semantic prior—\(S_{{weather}}\) enhances degradation textures in visible images, while \(S_{{scene}}\) strengthens object information in both infrared and visible features. This demonstrates the semantic prior's deep guidance on image features, highlighting critical regions to enable adaptive expert routing.
\subsubsection{Degradation-Aware Channel Attention Module (DCAM)}
We propose a degradation-aware module that leverages the semantic prior for channel modulation. Specifically, we utilize the semantic prior from DSPE, which encodes both weather-related $S_{weather}$ and scene-related $S_{scene}$ information to identify similar and distinguish different degradation scenarios.

To decompose the semantic prior into degradation prototypes, we first pass \(S_{prior}\) through a dimensionality reduction MLP layer $\Phi_{m}^{II}(\cdot)$ to produce a score vector \({s}_K \in \mathbb{R}^{K}\), where each entry corresponds to the activation score of one of the \(K\) degradation prototypes. It is formulated as:
\begin{align}
    {s}_K & = {\sigma}(\mathcal{N}_{layer}({\Phi_{m}^{II}}(\mathcal{P}_{avg}({S}_{prior})))),
\end{align}
where \(\mathcal{P}_{avg}(\cdot)\) denotes average pooling, and \(\sigma(\cdot)\) is the Sigmoid activation function.

The score vector \({s}_K\) is then used to represent the degradation scenario, expressed as a linear combination of the degradation prototypes:
\begin{align}
    D_{s} & = \sum\nolimits_{i=1}^{K} s_{K_i} \cdot \mathcal{P}_i,
\end{align}
where \(D_{s}\) denotes the specific degradation scenario, and \(\mathcal{P}_i\) denotes the \(i\)-th degradation prototype.

Then, we model the relationship between degradation prototypes and channel responses by representing each degradation prototype as a vector \(k_i \in \mathbb{R}^{C}\), where \(C\) is the number of channels in \(F_{in}\). Each vector encodes the response strength of the \(i\)-th prototype across the \(C\) channels. Consequently, the channel-wise weights can be obtained as:
\begin{align}
w_c = \sigma \left( \sum\nolimits_{i=1}^{K} s_{K_i} \cdot k_i \right),
\end{align}
where \(w_c \in \mathbb{R}^{C}\) represents the channel-wise weights.

To represent the degradation prototypes, we concatenate the \(K\) degradation prototype vectors \(k_i \in \mathbb{R}^{C}\) into a degradation prototype matrix \(W_{proto} \in \mathbb{R}^{K \times C}\). This matrix serves as the practical implementation of the degradation prototypes.
\begin{figure}[t!]
  \centering
  \includegraphics[width=\columnwidth]{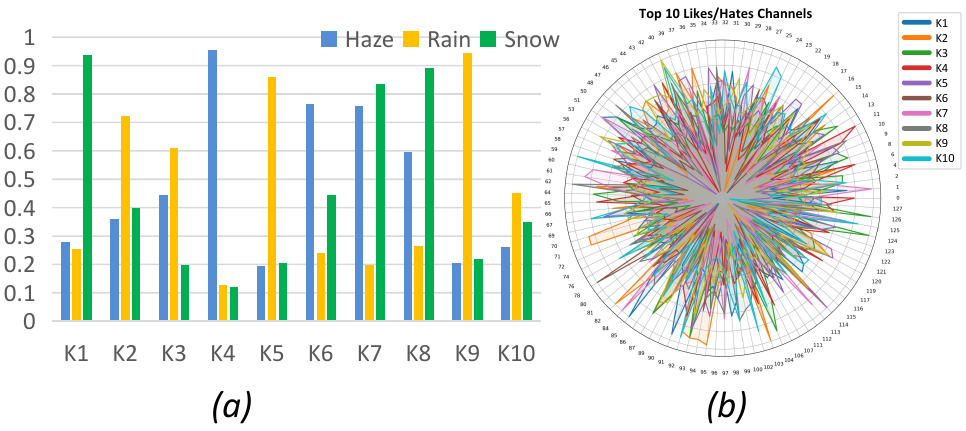}
  \small
  \caption{(a) Decomposition of semantic prior into degradation prototypes across different weather types. (b) Radar chart of top 10 activated (outward) and suppressed (inward) channels per degradation prototype.}
    \label{fig:dcam}
\end{figure}

To ensure distinct channel response patterns, prototype vectors are initialized orthogonally and normalized to the range $[-1, 1]$, representing activation strengths across channels. These vectors form the prototype matrix \(W_{proto}\), where each row corresponds to a distinct prototype vector. The normalized orthogonal matrix \(W_{proto}\) is then treated as a learnable parameter.

In summary, the semantic prior decomposition generates degradation prototypes that encode distinct channel response patterns, enabling degradation-aware channel modulation, formally expressed as:
\begin{align}
F_{dcam} =\mathcal{N}_{layer}(F_{in}) \odot \sigma\left( s_K W_{proto} \right) + F_{in},
\end{align}
where \(\odot\) denotes channel-wise element-wise multiplication.

\textbf{As shown in Fig.~\ref{fig:dcam} (a)}, under different weather conditions, the semantic prior is decomposed into a set of degradation prototypes, whose corresponding proportions display both clear distinctions and latent correlations across different scenarios. This demonstrates that the semantic prior effectively captures degradation-aware semantics — a high-level representation of degradtion senarios — which can be further factorized into a weighted composition of multiple degradation prototypes. \textbf{As illustrated in Fig.~\ref{fig:dcam} (b)}, our training strategy enables each prototype to learn diverse and distinct channel preference patterns. Such diversity improves the prototype mixture's expressiveness and adaptability to various degradation scenarios.
\begin{table*}[t!]
\centering
\small
\renewcommand{\arraystretch}{0.8}
\setlength{\tabcolsep}{1.9pt}
\resizebox{\textwidth}{!}{%
\begin{tabular}{lcccc|lcccc|lcccc|lcccc|cccc|cccc}
\toprule
\multicolumn{15}{c}{\textbf{Strategy I}} & \multicolumn{13}{c}{\textbf{Strategy II}} \\
\cmidrule(lr){1-15} \cmidrule(lr){16-28}
 & \multicolumn{4}{c}{\textbf{Haze}} & & \multicolumn{4}{c}{\textbf{Rain}} & & \multicolumn{4}{c}{\textbf{Snow}} 
 & & \multicolumn{4}{c}{\textbf{Haze}} & \multicolumn{4}{c}{\textbf{Rain}} & \multicolumn{4}{c}{\textbf{Snow}} \\
\cmidrule(lr){2-5} \cmidrule(lr){7-10} \cmidrule(lr){12-15} 
\cmidrule(lr){17-20} \cmidrule(lr){21-24} \cmidrule(lr){25-28}
\textbf{Method} & \textbf{PSNR↑} & \textbf{SSIM↑} & \textbf{NABF↓} & \textbf{MI↑} 
& \textbf{Method} & \textbf{PSNR↑} & \textbf{SSIM↑} & \textbf{NABF↓} & \textbf{MI↑} 
& \textbf{Method} & \textbf{PSNR↑} & \textbf{SSIM↑} & \textbf{NABF↓} & \textbf{MI↑}
& \textbf{Method} & \textbf{PSNR↑} & \textbf{SSIM↑} & \textbf{NABF↓} & \textbf{MI↑}
& \textbf{PSNR↑} & \textbf{SSIM↑} & \textbf{NABF↓} & \textbf{MI↑}
& \textbf{PSNR↑} & \textbf{SSIM↑} & \textbf{NABF↓} & \textbf{MI↑} \\
\midrule
\multicolumn{28}{c}{\textbf{MSRS Dataset}} \\
\midrule
\multicolumn{5}{l|}{\textbf{+ SegMiF}} & \multicolumn{5}{l|}{\textbf{+ SegMiF}} & \multicolumn{5}{l|}{\textbf{+ SegMiF}} 
& \multicolumn{5}{l|}{\textbf{+ SegMiF}} & \multicolumn{4}{l|}{} & \multicolumn{4}{l}{} \\  
DCMPNet & 16.769 & 1.019 & 0.0287 & 1.673 & DRSformer & 17.308 & 0.859 & 0.0719 & 1.722 & HDCWNet & 16.513 & 0.773 & 0.1149 & 1.594
& DTMR & 16.100 & 0.949 & 0.0384 & 1.523
& 15.672 & 0.579 & 0.1536 & 1.309
& 15.698 & 0.673 & 0.1175 & 1.778 \\
CasDyF-Net & 15.700 & 0.907 & 0.0203 & 1.779 & IDT & 16.515 & 0.608 & 0.1312 & 1.305 & InvDSNet & 16.051 & 0.612 & 0.1200 & 1.575
& MPMF-Net & 17.140 & 1.044 & 0.0208 & \underline{1.931}
& 17.535 & 1.038 & 0.0292 & 1.949
& 16.389 & 0.716 & 0.0926 & 1.693 \\
DehazeFormer & 17.051 & 1.046 & 0.0222 & \underline{2.183} & NeRD-Rain & 16.461 & 0.585 & 0.1355 & 1.289 & SnowFormer & 16.007 & 0.616 & 0.1224 & 1.606
& WGWS-Net & 16.241 & 0.943 & 0.0212 & 1.653
& 17.237 & 0.926 & 0.0563 & 1.843
& \underline{16.688} & 0.721 & 0.0915 & 1.797 \\
\midrule
\multicolumn{5}{l|}{\textbf{+ SwinFuse}} & \multicolumn{5}{l|}{\textbf{+ SwinFuse}} & \multicolumn{5}{l|}{\textbf{+ SwinFuse}} 
& \multicolumn{5}{l|}{\textbf{+ SwinFuse}} & \multicolumn{4}{l|}{} & \multicolumn{4}{l}{} \\  
DCMPNet & 15.631 & 0.442 & 0.0277 & 1.423 & DRSformer & 17.000 & 0.663 & \underline{0.0225} & 1.366 & HDCWNet & 15.518 & 0.518 & 0.0710 & 1.430
& DTMR & 15.854 & 0.740 & 0.0435 & 1.550
& 16.601 & 0.926 & 0.0716 & 1.395
& 15.855 & 0.803 & 0.0822 & 1.685 \\
CasDyF-Net & 16.592 & 0.743 & 0.0230 & 1.911 & IDT & 17.271 & 0.826 & 0.0369 & 1.246 & InvDSNet & 15.639 & 0.533 & 0.0755 & 1.403
& MPMF-Net & 16.254 & 0.551 & 0.0205 & 1.496
& 16.703 & 0.628 & \underline{0.0180} & 1.494
& 15.828 & 0.742 & 0.0649 & 1.612 \\
DehazeFormer & 15.558 & 0.364 & 0.0237 & 1.416 & NeRD-Rain & 17.261 & 0.830 & 0.0380 & 1.238 & SnowFormer & 15.445 & 0.499 & 0.0829 & 1.440
& WGWS-Net & 16.253 & 0.693 & 0.0224 & 1.463
& 16.800 & 0.647 & 0.0225 & 1.429
& 16.030 & \underline{0.810} & 0.0736 & 1.739 \\
\midrule
\multicolumn{5}{l|}{\textbf{+ TarDAL}} & \multicolumn{5}{l|}{\textbf{+ TarDAL}} & \multicolumn{5}{l|}{\textbf{+ TarDAL}} 
& \multicolumn{5}{l|}{\textbf{+ TarDAL}} & \multicolumn{4}{l|}{} & \multicolumn{4}{l}{} \\  
DCMPNet & 15.719 & 0.144 & 0.0226 & 0.612 & DRSformer & 16.146 & 0.223 & 0.0310 & 0.758 & HDCWNet & 15.839 & 0.236 & \underline{0.0578} & 0.799
& DTMR & 15.829 & 0.191 & 0.0320 & 0.805
& 16.256 & 0.343 & 0.0771 & 0.861
& 16.041 & 0.295 & 0.0739 & 0.941 \\
CasDyF-Net & 15.871 & 0.181 & 0.0267 & 0.705 & IDT & 16.255 & 0.279 & 0.0454 & 0.728 & InvDSNet & 15.884 & 0.254 & 0.0634 & 0.764
& MPMF-Net & 15.818 & 0.165 & 0.0231 & 0.674
& 16.020 & 0.194 & 0.0251 & 0.763
& 15.922 & 0.271 & \underline{0.0543} & 0.828 \\
DehazeFormer & 15.716 & 0.143 & 0.0212 & 0.614 & NeRD-Rain & 16.260 & 0.283 & 0.0463 & 0.733 & SnowFormer & 15.853 & 0.254 & 0.0648 & 0.794
& WGWS-Net & 15.821 & 0.170 & 0.0240 & 0.694
& 16.091 & 0.219 & 0.0292 & 0.772
& 16.014 & 0.307 & 0.0665 & 0.946 \\
\midrule
\multicolumn{5}{l|}{\textbf{+ MetaFusion}} & \multicolumn{5}{l|}{\textbf{+ MetaFusion}} & \multicolumn{5}{l|}{\textbf{+ MetaFusion}} 
& \multicolumn{5}{l|}{\textbf{+ MetaFusion}} & \multicolumn{4}{l|}{} & \multicolumn{4}{l}{} \\  
DCMPNet & 16.717 & 1.104 & 0.0590 & 1.394 & DRSformer & 17.058 & 0.917 & 0.1201 & 1.235 & HDCWNet & 16.210 & 0.831 & 0.1784 & 1.281
& DTMR & 16.067 & 1.138 & 0.0967 & 1.377
& 13.841 & 0.512 & 0.2342 & 0.986
& 14.405 & 0.704 & 0.2241 & 1.326 \\
CasDyF-Net & 15.530 & 1.040 & 0.0521 & 1.646 & IDT & 15.290 & 0.561 & 0.1928 & 0.903 & InvDSNet & 15.590 & 0.657 & 0.1924 & 1.183
& MPMF-Net & \underline{17.175} & 1.208 & 0.0499 & 1.478
& 17.604 & 1.132 & 0.0742 & 1.416
& 15.370 & 0.695 & 0.1797 & 1.197 \\
DehazeFormer & 16.940 & 1.149 & 0.0472 & 1.611 & NeRD-Rain & 15.226 & 0.535 & 0.1973 & 0.886 & SnowFormer & 15.668 & 0.669 & 0.1921 & 1.214
& WGWS-Net & 16.873 & 1.176 & 0.0507 & 1.460
& 17.162 & 0.991 & 0.1043 & 1.349
& 15.257 & 0.695 & 0.1735 & 1.315 \\
\midrule
\multicolumn{5}{l|}{\textbf{+ SAGE}} & \multicolumn{5}{l|}{\textbf{+ SAGE}} & \multicolumn{5}{l|}{\textbf{+ SAGE}} 
& \multicolumn{5}{l|}{\textbf{+ SAGE}} & \multicolumn{4}{l|}{} & \multicolumn{4}{l}{} \\  
DCMPNet & 16.589 & 1.171 & 0.0222 & 1.713 & DRSformer & \underline{17.964} & \underline{0.993} & 0.0704 & \underline{1.880} & HDCWNet & \underline{17.267} & \underline{0.897} & 0.1176 & 1.747
& DTMR & 15.722 & 1.141 & 0.0322 & 1.649
& 14.237 & 0.611 & 0.1721 & 1.542
& 14.818 & 0.797 & 0.1289 & 2.132 \\
CasDyF-Net & 13.957 & 1.025 & 0.0171 & 1.969 & IDT & 16.264 & 0.665 & 0.1373 & 1.490 & InvDSNet & 16.493 & 0.712 & 0.1243 & \underline{1.793}
& MPMF-Net & 17.105 & \underline{1.217} & \underline{0.0148} & 1.897
& \textbf{18.385} & \underline{1.189} & 0.0241 & \underline{2.178}
& 16.415 & 0.806 & 0.0981 & 2.019 \\
DehazeFormer & \underline{17.260} & \underline{1.231} & \underline{0.0157} & 2.005 & NeRD-Rain & 16.220 & 0.641 & 0.1418 & 1.481 & SnowFormer & 16.548 & 0.716 & 0.1261 & 1.752
& WGWS-Net & 15.951 & 1.141 & 0.0158 & 1.689
& \underline{18.203} & 1.056 & 0.0548 & 2.053
& 14.761 & 0.752 & 0.0979 & \underline{2.252} \\
\midrule
\textbf{Ours} & \textbf{18.325} & \textbf{1.302} & \textbf{0.0111} & \textbf{2.273} & \textbf{Ours} & \textbf{18.079} & \textbf{1.260} & \textbf{0.0131} & \textbf{2.406} & \textbf{Ours} & \textbf{17.528} & \textbf{1.245} & \textbf{0.0151} & \textbf{2.491}
& \textbf{Ours} & \textbf{18.325} & \textbf{1.302} & \textbf{0.0111} & \textbf{2.273}
& 18.079 & \textbf{1.260} & \textbf{0.0131} & \textbf{2.406}
& \textbf{17.528} & \textbf{1.245} & \textbf{0.0151} & \textbf{2.491} \\
\midrule
\multicolumn{28}{c}{\textbf{FMB Dataset}} \\
\midrule
\multicolumn{5}{l|}{\textbf{+ SegMiF}} & \multicolumn{5}{l|}{\textbf{+ SegMiF}} & \multicolumn{5}{l|}{\textbf{+ SegMiF}} 
& \multicolumn{5}{l|}{\textbf{+ SegMiF}} & \multicolumn{4}{l|}{} & \multicolumn{4}{l}{} \\  
DCMPNet & 12.267 & 1.041 & 0.1026 & 2.329 & DRSformer & 13.440 & 1.051 & \underline{0.0288} & 2.550 & HDCWNet & 12.587 & 0.600 & 0.1267 & 1.781
& DTMR & 12.420 & 1.016 & 0.0486 & 2.292
& 12.566 & 0.790 & 0.0804 & 1.897
& 12.416 & 0.778 & 0.0771 & 1.998 \\
CasDyF-Net & 13.325 & 1.096 & 0.0355 & 2.458 & IDT & 13.457 & 0.960 & 0.0423 & 2.327 & InvDSNet & 12.332 & 0.533 & \underline{0.1097} & 1.781
& MPMF-Net & 13.383 & 1.229 & 0.0168 & 2.267
& 12.990 & 1.033 & 0.0281 & 2.504
& 13.085 & 0.867 & 0.0657 & 2.355 \\
DehazeFormer & 12.486 & 1.128 & \underline{0.0193} & 2.540 & NeRD-Rain & 13.482 & 0.968 & 0.0408 & 2.332 & SnowFormer & 12.457 & 0.553 & 0.1103 & \underline{2.189}
& WGWS-Net & 12.800 & 1.179 & \underline{0.0134} & 2.273
& 13.365 & 1.096 & \underline{0.0199} & 2.677
& 13.608 & 0.862 & \underline{0.0592} & 2.292 \\
\midrule
\multicolumn{5}{l|}{\textbf{+ SwinFuse}} & \multicolumn{5}{l|}{\textbf{+ SwinFuse}} & \multicolumn{5}{l|}{\textbf{+ SwinFuse}} 
& \multicolumn{5}{l|}{\textbf{+ SwinFuse}} & \multicolumn{4}{l|}{} & \multicolumn{4}{l}{} \\  
DCMPNet & 12.448 & 1.046 & 0.0424 & 2.489 & DRSformer & \underline{15.855} & 1.249 & 0.0384 & \underline{2.821} & HDCWNet & 13.462 & 0.732 & 0.1493 & 1.965
& DTMR & 13.650 & 1.164 & 0.0528 & 2.427
& 14.484 & 1.007 & 0.0882 & 2.202
& 14.015 & 0.979 & 0.0880 & 2.235 \\
CasDyF-Net & 14.259 & 1.235 & 0.1242 & \underline{2.788} & IDT & 15.851 & 1.179 & 0.0506 & 2.675 & InvDSNet & 13.051 & 0.635 & 0.1382 & 1.970
& MPMF-Net & 13.892 & 1.306 & 0.0420 & 2.455
& 14.958 & 1.217 & 0.0342 & 2.723
& 14.295 & 1.015 & 0.0949 & \underline{2.635} \\
DehazeFormer & 13.641 & 1.223 & 0.0285 & \textbf{3.508} & NeRD-Rain & 15.851 & 1.183 & 0.0494 & 2.682 & SnowFormer & 13.158 & 0.641 & 0.1456 & 2.189
& WGWS-Net & 14.387 & 1.248 & 0.0299 & 2.240
& 15.071 & 1.183 & 0.0353 & 2.668
& 13.276 & 0.823 & 0.1082 & 2.312 \\
\midrule
\multicolumn{5}{l|}{\textbf{+ TarDAL}} & \multicolumn{5}{l|}{\textbf{+ TarDAL}} & \multicolumn{5}{l|}{\textbf{+ TarDAL}} 
& \multicolumn{5}{l|}{\textbf{+ TarDAL}} & \multicolumn{4}{l|}{} & \multicolumn{4}{l}{} \\  
DCMPNet & 10.719 & 0.484 & 0.1028 & 1.995 & DRSformer & 12.291 & 0.674 & 0.1156 & 2.056 & HDCWNet & 12.130 & 0.475 & 0.1657 & 1.556
& DTMR & 11.249 & 0.583 & 0.1134 & 2.137
& 11.666 & 0.512 & 0.1423 & 1.710
& 11.556 & 0.477 & 0.1333 & 1.723 \\
CasDyF-Net & 12.727 & 0.850 & 0.1702 & 2.176 & IDT & 12.402 & 0.652 & 0.1267 & 1.898 & InvDSNet & 11.853 & 0.411 & 0.1632 & 1.484
& MPMF-Net & 13.374 & 1.027 & 0.1023 & 2.141
& 11.791 & 0.628 & 0.1091 & 2.120
& 12.444 & 0.633 & 0.1440 & 1.941 \\
DehazeFormer & 11.448 & 0.732 & 0.1023 & 2.190 & NeRD-Rain & 12.412 & 0.654 & 0.1249 & 1.951 & SnowFormer & 12.002 & 0.429 & 0.1667 & 1.549
& WGWS-Net & 12.344 & 0.885 & 0.1001 & 2.180
& 12.175 & 0.711 & 0.1008 & 2.279
& 12.976 & 0.649 & 0.1386 & 1.889 \\
\midrule
\multicolumn{5}{l|}{\textbf{+ MetaFusion}} & \multicolumn{5}{l|}{\textbf{+ MetaFusion}} & \multicolumn{5}{l|}{\textbf{+ MetaFusion}} 
& \multicolumn{5}{l|}{\textbf{+ MetaFusion}} & \multicolumn{4}{l|}{} & \multicolumn{4}{l}{} \\  
DCMPNet & 14.204 & 1.177 & 0.1182 & 2.002 & DRSformer & 15.533 & 1.139 & 0.0840 & 2.080 & HDCWNet & 13.398 & 0.627 & 0.2101 & 1.489
& DTMR & 14.714 & 1.197 & 0.1305 & 2.007
& 14.563 & 0.848 & 0.1689 & 1.577
& 14.208 & 0.916 & 0.1706 & 1.714 \\
CasDyF-Net & 13.195 & 1.048 & 0.2201 & 2.252 & IDT & 15.214 & 1.001 & 0.0993 & 1.890 & InvDSNet & 13.059 & 0.571 & 0.2045 & 1.412
& MPMF-Net & 12.157 & 1.197 & 0.0733 & 2.021
& \underline{15.702} & 1.155 & 0.0881 & 2.003
& 13.843 & 0.877 & 0.1525 & 1.934 \\
DehazeFormer & \underline{14.674} & 1.282 & 0.0855 & 2.367 & NeRD-Rain & 15.226 & 1.002 & 0.0975 & 1.892 & SnowFormer & 13.205 & 0.591 & 0.2099 & 1.517
& WGWS-Net & \underline{14.809} & 1.296 & 0.0630 & 2.085
& 15.683 & 1.230 & 0.0740 & 2.200
& 12.233 & 0.816 & 0.1275 & 1.796 \\
\midrule
\multicolumn{5}{l|}{\textbf{+ SAGE}} & \multicolumn{5}{l|}{\textbf{+ SAGE}} & \multicolumn{5}{l|}{\textbf{+ SAGE}} 
& \multicolumn{5}{l|}{\textbf{+ SAGE}} & \multicolumn{4}{l|}{} & \multicolumn{4}{l}{} \\  
DCMPNet & 14.592 & 1.338 & 0.0407 & 2.602 & DRSformer & 15.164 & \underline{1.266} & 0.0312 & 2.730 & HDCWNet & \underline{13.594} & \underline{0.808} & 0.1268 & 1.915
& DTMR & 14.670 & 1.344 & 0.0514 & \underline{2.581}
& 14.681 & 1.007 & 0.0956 & 2.129
& \underline{14.379} & \underline{1.074} & 0.0878 & 2.246 \\
CasDyF-Net & 12.963 & 1.212 & 0.0962 & 1.878 & IDT & 14.801 & 1.137 & 0.0443 & 2.520 & InvDSNet & 13.305 & 0.710 & 0.1166 & 1.869
& MPMF-Net & 11.196 & 1.293 & 0.0234 & 2.467
& 15.555 & 1.316 & 0.0294 & 2.687
& 13.513 & 1.049 & 0.0693 & 2.538 \\
DehazeFormer & 14.223 & \underline{1.380} & 0.0225 & 2.775 & NeRD-Rain & 14.808 & 1.144 & 0.0430 & 2.525 & SnowFormer & 13.370 & 0.735 & 0.1206 & 2.044
& WGWS-Net & 13.412 & \underline{1.370} & 0.0176 & 2.562
& 15.378 & \underline{1.353} & 0.0201 & \underline{2.884}
& 11.541 & 0.983 & 0.0604 & 2.367 \\
\midrule
\textbf{Ours} & \textbf{15.888} & \textbf{1.398} & \textbf{0.0103} & 2.756 
& \textbf{Ours} & \textbf{16.250} & \textbf{1.399} & \textbf{0.0106} & \textbf{3.006} 
& \textbf{Ours} & \textbf{16.231} & \textbf{1.385} & \textbf{0.0114} & \textbf{2.977}
& \textbf{Ours} & \textbf{15.888} & \textbf{1.398} & \textbf{0.0103} & \textbf{2.756}
& \textbf{16.250} & \textbf{1.399} & \textbf{0.0106} & \textbf{3.006}
& \textbf{16.231} & \textbf{1.385} & \textbf{0.0114} & \textbf{2.977} \\
\bottomrule
\end{tabular}
}
\caption{Quantitative comparison under Strategy I and II on Hazy, rainy and snowy MSRS and FMB datasets.}
\label{tab:Quantitative}
\end{table*}
\subsubsection{Degradation-Aware Mixture-of-Experts (DMoE)}
To implement a degradation-aware MoE network, we establish cross-modal interactions between the semantic prior \( {S}_{{prior}} \) and the DCAM-weighted feature maps \( {F}_{dcam} \), generating expert activation scores.
\begin{figure}[t]
  \centering
  \includegraphics[width=1\columnwidth]{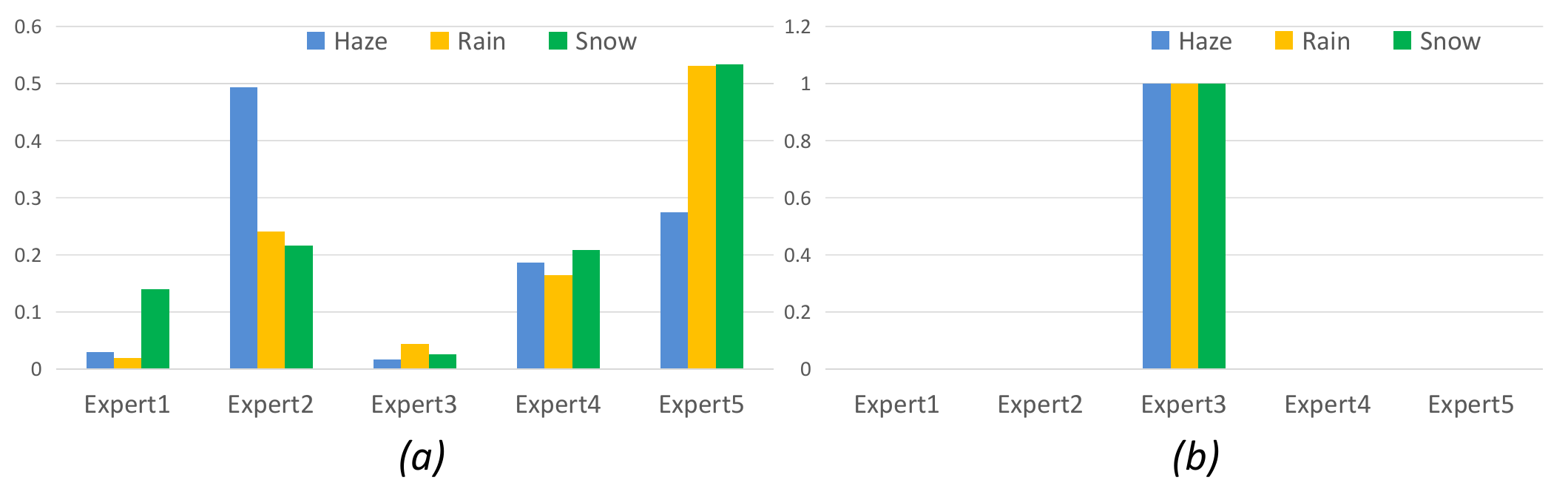}
  \caption{
    Expert activation patterns comparison on test sets: (a) Full model; (b) Model without DCAM and DMoE, lacking semantic prior guidance.
    }
    \label{fig:expert}
\end{figure}

We use a cross-attention mechanism to interact the semantic prior \( {S}_{{prior}} \) with image features, emphasizing those features related to the degradation scenario. It is expressed as:
\begin{align}
    {Q} &= {F}_{{dcam}} {W}_{Q},{K} = {S}_{{prior}} {W}_{K},{V} = {S}_{{prior}} {W}_{V},\\
    \hat{{a}} &= {\mathcal{FL}}({Softmax}\left(\frac{{Q}{K}^\top}{\sqrt{d_k}}\right){V}+{F}_{{dcam}}),
\end{align}
where \(\mathcal{FL}(\cdot)\) denotes the flattening operation.
The cross-attention output is then processed through a dual-branch layer to reduce dimensionality:
{\small
\begin{equation}
     F_{b}^{I} = {\phi}\left( {\mathcal{N}_{layer}}(\mathcal{T}( {\hat{a}})) \right), 
     F_{b}^{II} = {\mathcal{P}_{avg}}\left( {\phi}(\mathcal{N}_{layer}( {\hat{a}})) \right),
\end{equation}
}
where \( \phi(\cdot) \) denotes the GELU activation function, \( \mathcal{T}(\cdot) \) represents a linear transformation layer used for dimensionality reduction, and \( F_{b}^{I} \), \( F_{b}^{II} \) are then passed through a convolutional layer \( \mathcal{C}(\cdot) \) for channel reduction. This is followed by a softmax operation to compute the expert activation scores \( w \in \mathbb{R}^{1 \times N} \), where \( N \) denotes the number of experts:
\begin{equation}
    w = \mathrm{Softmax}\left( \mathcal{C}\left( \mathrm{Cat}\left(F_{b}^{I}, F_{b}^{II}\right) \right) \right).
\end{equation}

Then, \( F_{dcam} \) is processed in parallel by a set of \( N = 5 \) experts, each sharing an identical lightweight architecture. Specifically, each expert comprises a \( 3 \times 3 \) convolutional layer \( \mathcal{C}_{3\times3}(\cdot) \) followed by a \( 1 \times 1 \) convolution \( \mathcal{C}_{1\times1}(\cdot) \). The output of this expert ensemble is formulated as:
\begin{align}
    {E}_i &= {\mathcal{C}}_{1\times1}\left( {\mathcal{C}}_{3\times3}({F}_{dcam}) \right), \\
    {F}_{dmoe} &= {\phi}\left( {\mathcal{N}_{batch}}\left( \sum\nolimits_{i=1}^{N} {w}_i {E}_i \right) \right),
\end{align}
where \( E_i \) denotes the output of the \( i \)-th expert, \( w_i \) indicates its corresponding activation weight, and $\mathcal{N}_{batch}(\cdot)$ represents batch normalization.

\textbf{As illustrated in Fig.~\ref{fig:expert} (a)}, our semantic prior-guided routing activates all experts with diverse assignments across weather conditions, indicating effective degradation-aware selection. In contrast, \textbf{Fig.~\ref{fig:expert} (b)} shows that removing DCAM and DMoE causes expert collapse, with a single expert dominating, reflecting a loss of degradation-awareness. These results highlight the crucial role of semantic priors in enabling dynamic expert selection.

\subsubsection{Decoder}
The enhanced feature ${F}_{dmoe}$ is then fed into a CNN-based decoder $\mathcal{D}(\cdot)$ for image reconstruction, as formulated below:
\begin{equation}
    {I}_f = \mathcal{D}({F}_{dmoe}).
\end{equation}
\subsection{Loss Functions}
We design loss functions based on the framework from prior work~\cite{zhang2024mrfs}. To address the degradation, we adapt it to the fusion-based degradation removal task. Specifically, we use the original undegraded visible image \( \tilde{{I}}_{{vi}} \) and \( I_{{ir}} \) as ground-truth.The overall image fusion loss \( L_{{fusion}} \) is the sum of the integration $L_{{inte}}$ and color consistency losses $L_{color}$:
\begin{equation}
    L_{{fusion}} = L_{{inte}} + L_{{color}}.
\end{equation}
\subsubsection{Integration Loss}
To preserve texture and edge details, we define the integration loss \( L_{{inte}} \), which ensures the fused image matches the pixel-wise intensity and gradient maxima of the source images. It is formulated as:
{
\small
\begin{equation}
    L_{inte}\!= \! \left\|\! \nabla\! I_f\!-\! \max(\nabla\! \tilde{I}_{vi},\! \nabla I_{ir})\! \right\|_1\! +\! \left\|\! I_f\! -\! \max(\tilde{I}_{vi},\! I_{ir})\! \right\|_1\! ,
\end{equation}
}
where \( \nabla I \) represents the image gradient, and \( \max(\cdot) \) denotes the pixel-wise maximum operation.
\subsubsection{Color Consistency Loss}
To align the chrominance with the visible image, we introduce a color consistency loss \( L_{{color}} \), which is defined as:
\begin{equation}
    L_{{color}} = \left\| C_b^{f} - C_b^{\tilde{{vi}}} \right\|_1 + \left\| C_r^{f} - C_r^{\tilde{{vi}}} \right\|_1,
\end{equation}
where \( C_b \) and \( C_r \) are the chrominance components in the YCbCr color space. Superscripts \( f \) and \( \tilde{{vi}} \) refer to the fused and clean visible images, respectively.
\begin{figure*}[t]
  \centering
  \includegraphics[width=1\textwidth]{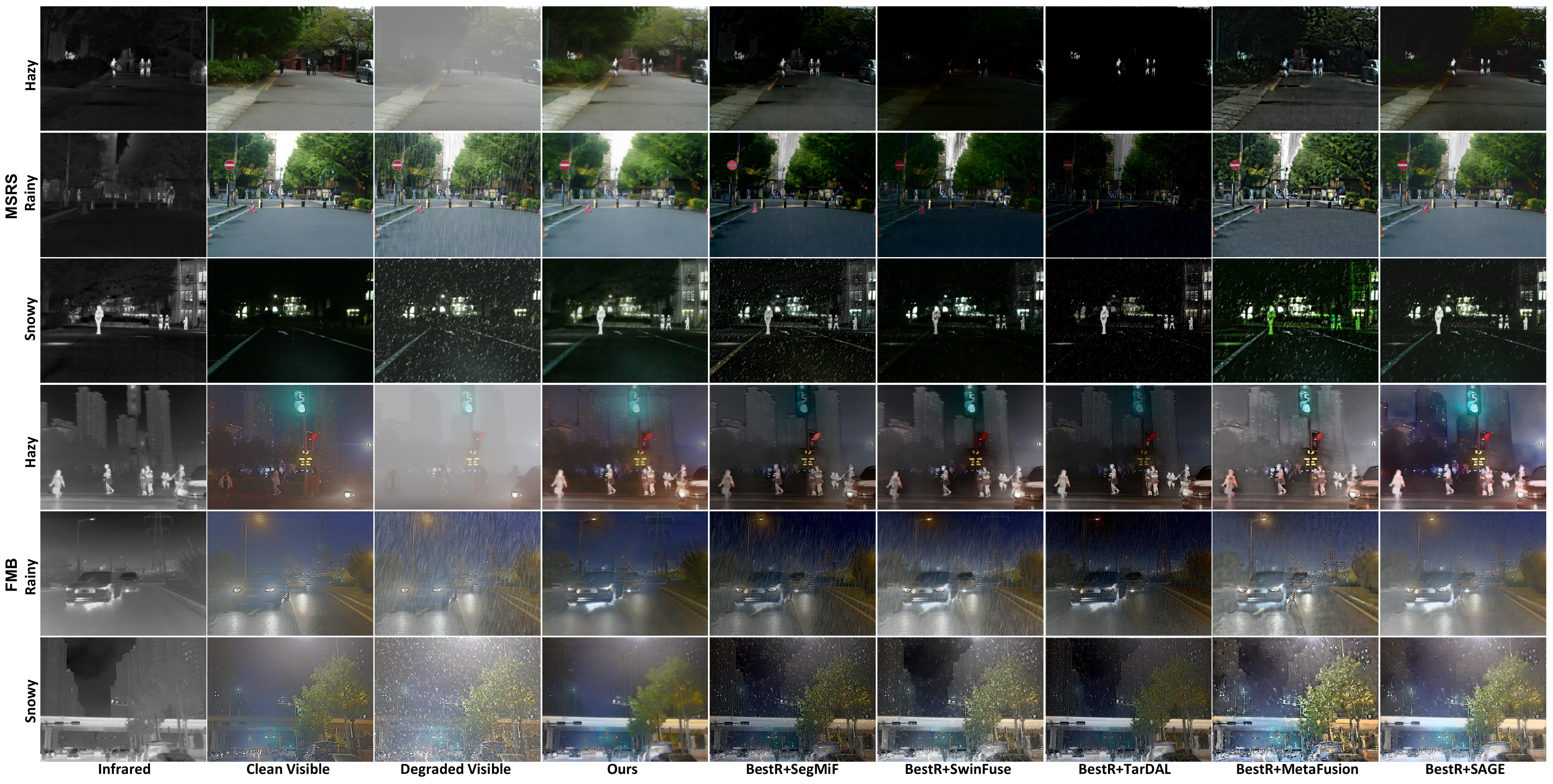}
  \small
  \caption{Qualitative comparison on degraded MSRS and FMB datasets. Rows 1-3: MSRS dataset, organized by weather conditions. Rows 4-6: FMB dataset, also grouped by weather.}
    \label{fig:qualitative}
\end{figure*}
\section{Experiments}
\subsection{Experimental Configurations}
\subsubsection{Datasets}
The datasets used in our experiments are derived from two widely adopted IVF benchmarks: MSRS~\cite{msrs} and FMB~\cite{liu2023segmif},  both providing clean and well-aligned visible-infrared image pairs. MSRS contains 1,524 pairs ($480 \times 640$), split into 1,163 for training and 361 for testing. FMB includes 1,500 aligned pairs ($600 \times 800$), with 1,220 for training and 280 for testing.

To simulate adverse weather conditions, we generated three degraded versions of the full MSRS and FMB datasets by applying haze, rain, and snow effects separately to all visible images. Specifically, haze was synthesized using the Depth-Anything model~\cite{depthanything} and the atmospheric scattering model~\cite{zhang2017hazerd}; rain was produced by combining random noise and motion blur; and snow was generated using the imgaug library~\cite{imgaug}. This process resulted in three complete degraded variants of the original datasets, leading to a total of 9,072 registered image pairs (7,149 for training and 1,923 for testing), evenly distributed among the three degradation types (1:1:1).
\subsubsection{Implementation Details}
We set the initial learning rate to $6 \times 10^{-5}$, with a batch size of 15, and train the model for 1,000 epochs in total. The learning rate is scheduled using a combination of warm-up and polynomial decay, and we employ the Adam optimizer for optimization. All experiments are conducted on NVIDIA GeForce RTX 4090 GPU with 24GB memory and the AMD Ryzen 9 7950X 16-Core Processor CPU.
\subsection{Quantitative Comparison}
To validate the superiority of our degradation-aware fusion framework across multiple degradation scenarios, we compare it with two strategies: \textbf{Strategy I}, which applies dedicated restoration models for each degradation type (haze, rain, snow) followed by baseline fusion; and \textbf{Strategy II}, which uses all-in-one restoration models to preprocess degraded inputs before baseline fusion. \emph{In contrast, our method directly fuses original degraded inputs without separate restoration, yielding more effective and streamlined performance.}

We conduct evaluations on 1,923 synthetically degraded image pairs from the MSRS and FMB datasets, benchmarking against five SOTA fusion methods (SAGE~\cite{wu2025sage}, SegMiF~\cite{liu2023segmif}, MetaFusion~\cite{zhao2023metafusion}, TarDAL~\cite{liu2022tardal}, SwinFuse~\cite{wang2022swinfuse}) combined with restoration networks. These include dedicated restoration models for dehazing (DCMPNet~\cite{DCMP2024}, CasDyF-Net~\cite{casdyf2024}, DehazeFormer~\cite{dehazeformer2023vision}), deraining (DRSformer~\cite{drsformer_Chen_2023_CVPR}, IDT~\cite{IDT_xiao2022image}, NeRD-Rain~\cite{NeRD-Rain}), desnowing (HDCWNet~\cite{hdcw2021all}, InvDSNet~\cite{InvDSNet_quan2023image}, SnowFormer~\cite{SnowFormer_chen2022snowformer}), and all-in-one restoration models (DTMR~\cite{dtmr_Patil_2023_ICCV}, MPMF-Net~\cite{mpmfnet2025multi}, WGWS-Net~\cite{wgwsnet_zhu2023Weather}), all paired with fixed fusion modules. Performance is evaluated by PSNR, SSIM, Nabf, and MI metrics.

Quantitative results under both strategies are summarized in Table~\ref{tab:Quantitative}, where \textbf{↑} and \textbf{↓} denote preferable higher or lower values, and \textbf{bold} and \underline{underline} highlight the best and second-best results, respectively. Our method achieves the best performance in most cases, as evidenced by the quantitative metrics.
\subsection{Qualitative Comparison}
To identify the optimal restoration model for each baseline fusion method, we select the restoration model achieving the highest PSNR for each degradation type and dataset based on Table~\ref{tab:Quantitative}, and denote it as \textbf{BestR}. Using BestR, we construct fusion pipelines for baseline methods and qualitatively compare them with our approach, as illustrated in Fig.~\ref{fig:qualitative}. Our method consistently delivers clearer, higher-quality fused images across diverse weather conditions and datasets, outperforming SOTA techniques.
\subsection{Ablation Study}
\begin{table}[t!]
\centering
\small
\renewcommand{\arraystretch}{0.8}
\setlength{\tabcolsep}{2pt} 
\begin{tabular}{cccccc}
\toprule
\textbf{DCAM} & \textbf{DMoE} & \textbf{PSNR↑}           & \textbf{SSIM↑}          & \textbf{NABF↓}           & \textbf{MI↑}             \\
\cmidrule{1-6}
$\times$      & $\times$           & 16.265 & 1.144 & 0.0142                   & 2.006                    \\
$\times$      & $\checkmark$       & 15.350                   & 1.083                   & 0.0148                   & \underline{2.159}   \\
$\checkmark$  & $\times$           & \underline{17.009}                   & \underline{1.194}                   & \underline{0.0135}  & 2.101                    \\
$\checkmark$  & $\checkmark$       & \textbf{17.977}  & \textbf{1.269}  & \textbf{0.0131} & \textbf{2.390}  \\
\bottomrule
\end{tabular}
\caption{Average ablation results across haze, rain, and snow on the MSRS dataset.}
\label{tab:ablation}
\end{table}
Ablation studies in Table~\ref{tab:ablation} validate the effectiveness of DCAM and DMoE, with optimal results only when both are used. Removing DCAM disables channel-wise modulation, while removing DMoE makes routing purely image-driven, leading to performance drops (first row) and expert collapse—where only one expert remains active, as visualized in Fig.~\ref{fig:expert}(b). These results underscore the necessity of semantic prior guidance for degradation-aware routing and overall performance.
\section{Conclusion}
In this paper, we propose MdaIF, a robust one-stop multi-degradation-aware image fusion framework guided by the semantic prior from a pre-trained VLM. Unlike general IVF methods that overlook visible image degradation, MdaIF adapts fusion via two core modules: DCAM, which uses degradation prototype decomposition to modulate channel-wise features, and DMoE, enabling adaptive expert routing under adverse conditions. Compared to fixed-architecture IVF and prompt-guided fusion methods—which rely on either rigid networks or ground-truth degradation labels—our framework eliminates such dependencies by leveraging the semantic prior. MdaIF consistently outperforms SOTA methods across varied degradations, demonstrating strong effectiveness and generalization.

\section{Acknowledgments}
This research is partially supported by the National Natural Science Foundation of
China under Grant 62206321; and the Science and Technology Program of Hunan Province under Grant 2024RC3108.
\bibliography{aaai2026}
\end{document}